\documentclass{article}

\usepackage{arxiv}

\usepackage{amsmath,amssymb}
\usepackage{mathtools}
\usepackage{wasysym}
\usepackage[vlined,linesnumbered,ruled,titlenotnumbered]
{algorithm2e}

\usepackage{tikz}
\usepackage{tkz-euclide}
\usepackage{xspace}
\usepackage{epsfig}
\usepackage{color}
\usepackage{url}
\usepackage{graphicx,dsfont}
\usepackage{tikz}
\usepackage{float}
\usepackage{longtable}
\usepackage{booktabs}
\usepackage{makecell}
\usepackage{multicol}
\usepackage{multirow}
\usepackage{colortbl}
\usepackage{tikz}
\usepackage{csquotes}

\definecolor{uniform1}{HTML}{1B9E77}
\definecolor{uniform2}{HTML}{D95F02}
\definecolor{uniform3}{HTML}{7570B3}
\definecolor{poisson}{HTML}{E7298A}

\newcommand{\hide}[1]{}

\usepackage{xcolor}

\definecolor{jakobcolor}{rgb}{0.2,0.6,0.6}
\definecolor{frankcolor}{rgb}{0,0.4,0.8}
\definecolor{todocolor}{rgb}{0.9,0.1,0.1}
\definecolor{changedcolor}{rgb}{0.42,0.27,0.57}
\definecolor{addedcolor}{rgb}{0.867,0.176,0.361}

\newcommand{\changed}[1]{\nbc{CHANGED}{#1}{changedcolor}}
\newcommand{\added}[1]{\nbc{ADDED}{#1}{addedcolor}}

\newcommand{\redacted}[1]{\emph{[anonymized for submission]}}
\renewcommand{\redacted}[1]{#1}

\renewcommand{\changed}[1]{#1}
\renewcommand{\added}[1]{#1}

\title{Exploring the Feature Space of TSP Instances Using Quality Diversity}

\author{
  Jakob Bossek \\
  Statistics and Optimization\\
  University of M\"unster \\
  M\"unster, Germany \\
  \texttt{bossek@wi.uni-muenster.de} \\
   \And
  Frank Neumann \\
  Optimisation and Logistics\\
  The University of Adelaide\\
  Adelaide, Australia \\
  \texttt{frank.neumann@adelaide.edu.au}
}

\begin{document}
\maketitle

\begin{abstract}
  Generating instances of different properties is key to algorithm selection methods that differentiate between the performance of different solvers for a given combinatorial optimization problem. A wide range of methods using evolutionary computation techniques has been introduced in recent years. With this paper, we contribute to this area of research by providing a new approach based on quality diversity (QD) that is able to explore the whole feature space. QD algorithms allow to create solutions of high quality within a given feature space by splitting it up into boxes and improving solution quality within each box. We use our QD approach for the generation of TSP instances to visualize and analyze the variety of instances differentiating various TSP solvers and compare it to instances generated by established approaches from the literature.
\end{abstract}


\maketitle

\section{Introduction}

Evolutionary algorithms have been used for a wide range of different tasks. This includes solving classical $\mathcal{NP}$-hard optimization problems, optimizing the design for complex engineering problems~\cite{FLEMING20021223}, algorithms for machine learning~\cite{DBLP:journals/csur/TelikaniTBG22}, and approaches in the area of algorithm selection and configuration~\cite{Kerschke2019aas}.

In this paper, we focus on a crucial problem in the context of algorithm selection, namely the design of problem instances where two algorithms show a significantly different behaviour. We consider instances for the classical Euclidean traveling salesperson problem~(TSP) for which a wide range of algorithms has been introduced over the last 50 years~\cite{10.5555/1374811}. The problem is to find a round-trip tour of minimal costs through $n$ cities where the cities are given by coordinates in the 2D Euclidean space and pairwise distances correspond to the Euclidean distances of the points.

The complementary ability of algorithms for solving the TSP has been subjected to a wide range of studies. Center of these investigations is the characterization of features of TSP instances where a considered set of algorithms shows a significantly different behaviour. Studies have considered simplified variants of local search~\cite{smithmiles2010}, approximation algorithms~\cite{nallaperuma2013}, and state-of-the-art inexact TSP solvers such as EAX and LKH~\cite{BT2016EvolvingInstances,BT2016UnderstandingCharacteristics,kerschke2018}.

Evolutionary diversity optimization~(EDO) has recently been applied using various diversity measures to create diverse sets of images and TSP instances~\cite{alexander2017evolution,neumann2018discrepancy,neumann2019evolutionary}. For TSP instances, it has been shown that such approaches are able to create TSP instances with desired properties (such as good or bad performance behaviour) with a much wider range in the feature space~\cite{doi:10.1162/evcoa00274}.
Furthermore, recent investigations have used and analyzed EDO methods to create diverse sets of high quality solutions for the TSP and other permutation problems~\cite{viet2020evolving,DoGN021,NikfarjamBN021a,NikfarjamB0N21b}.

Quality diversity~(QD)~\cite{PughSS16} follows a similar goal as EDO. Algorithms explore a feature space of possible solutions to a given problem. Simple approaches, e.~g.,Map-Elites, partition the feature space into boxes and store for each box the best solution that has been found for its range of feature combinations.
QD algorithms have mainly been applied in the context of robotics and games~\cite{CullyM13,ZardiniZZIF21}. 
In the context of combinatorial optimization, a QD approach has recently been used for the traveling thief problem and it has been shown that significantly better solutions can be obtained~\cite{NNN2021qdttp}.

\changed{In this paper, we use for the first time the quality diversity approach in the context of instance generation}. We use a Map-Elites approach which for a given set of features divides the feature space into boxes and stores for each box TSP instances that show the maximum performance difference for two given algorithms.

Using this approach, we explore the feature space and optimize instances within each box with respect \changed{to solver performance}.
This allows to get an overview on differences of instances in the feature space in terms of their difficulties for a given set of TSP solvers.
In our experiments we evolve instances for the insertion heuristics Farthest-Insertion and Nearest-Insertion as well as for state-of-the-art exact TSP solvers EAX and LKH on different two-dimensional feature spaces. We compare our QD approach against \changed{classical $(\mu+1)$-EAs and EDO-tailored $(\mu+1)$-EAs modified to store their "footprint" in a similar way. Our experimental investigations show that with the proposed QD approach (a) in most cases a much wider range of feature combinations is explored during the evolutionary process and (b) the objective scores of the evolved instances are either comparable or even better in all cases}. Put together the QD approach is capable of generating a huge amount of interesting instances in a single run.

The paper is organized as follows. In Section~\ref{sec:problem} we describe the task of instance generation in detail, discuss challenges and review approaches from the literature. In Section~\ref{sec:method} we introduce our quality diversity approach. Experimental studies on simple and state-of-the-art TSP heuristics follow in Section~\ref{sec:experiments_cheap} and Section~\ref{sec:experiments_expensive} respectively. We conclude the paper in Section~\ref{sec:conclusion}.

\section{Problem formulation and history}
\label{sec:problem}

Let $\mathcal{A}$ be a portfolio of algorithms which show complementary performance on a set of problem instances $\mathcal{I}$. Further, let $F:\mathcal{I} \to \mathcal{F} \subseteq \mathbb{R}^p, p \geq 1$ be a \emph{feature-mapping} which maps instances to a $p$-dimensional real-valued numeric vector in the \emph{feature-space} $\mathcal{F}$ that ideally captures the characteristics of the instance which makes it easy or hard for the algorithms in the portfolio. In the context of the Euclidean TSP, features are for example the share of points on the convex-hull of the point coordinates or structural properties of transformations of the original instance, e.~g., statistics on the weak/strong connected components of a $k$-nearest neighbor graph~($k$-NNG). The idea of \emph{(per-instance) algorithm selection}~(AS) is to learn an algorithm selector $S : \mathcal{F} \to \mathcal{A}$ which -- for a previously unseen instance -- predicts the algorithm which will likely perform best on it based on its cheap-to-evaluate features only~\cite{Rice1976ASProblem,Kerschke2019aas}.
\begin{algorithm}[t]
	\SetKwInOut{Input}{input}
    \Input{Objective $f$, population size $\mu$.}
    Initialize population $P$ of $\mu$ instances at random\;
	\While{Stopping condition not met}{
        Select random instance $I \in P$\;
        Generate $I'$ by mutating $I$\;
        \If{$f(I')$ is not worse than $f(I)$}{
            Replace $I$ with $I'$ in $P$\;
        }
    }
    \Return{Best instance from $P$}\;
    \caption{Classical $(\mu+1)$~EA for instance generation.}
    \label{alg:ea_evolver}
\end{algorithm}

\begin{figure}
    \centering
    \includegraphics[width=0.95\columnwidth]{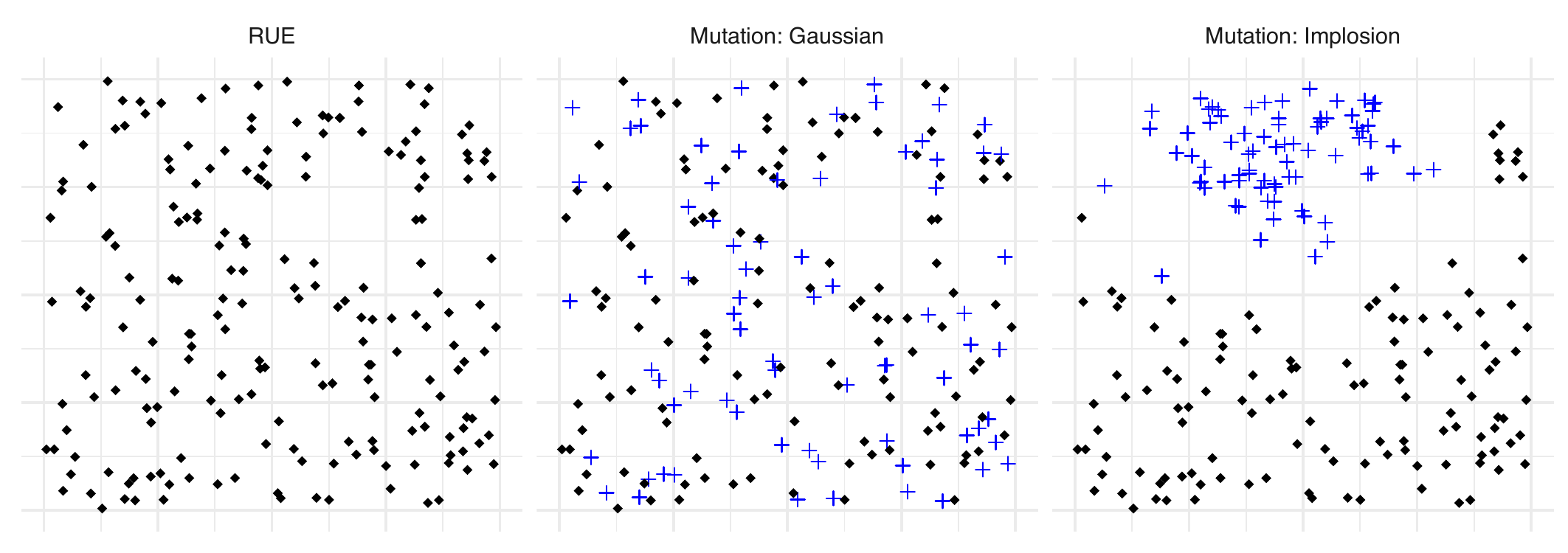}
    \caption{Random uniform Euclidean instance (left) with Gaussian mutation~(center) and implosion mutation~(right) applied to it. Blue points are affected by the mutation.}
    \label{fig:mutation_operators}
\end{figure}

In order to understand the strengths and weaknesses of algorithms and to be able to learn reasonable selectors, the instance set $\mathcal{I}$ should ideally be diverse on two different levels: (D1) diverse/complementary with respect to algorithm performance and (D2) diverse with respect to the structure of the instances which is captured by the coverage of the feature-space. Finding such a set of instances is challenging~\cite{smithmiles2010}. \changed{Since the availability of real-world instances for research is limited}, a systematic generation process is desirable. In this paper we focus on the problem of evolving a large set of diverse instances that satisfy both diversity requirements. 
For the TSP, early approaches focused on diversity in the performance space (D1). To this end Mersmann et al.~\cite{Mersmann13-TSPmeta} introduced a $(\mu+1)$~EA to generate instances which are hard/easy for an algorithm $A$ by minimizing/maximizing the gap to a known optimum; they evolved small instances for classical approximation algorithms and local-search algorithms. Subsequent works evolved instances which are easy for one algorithm $A_1$ and hard(er) for another algorithm $A_2$ by minimizing the ratio of performances measured by the \emph{penalized average runtime~(PAR~\cite{bischl2016})} for state-of-the-art inexact TSP solvers EAX and LKH~\cite{BT2016EvolvingInstances,BT2016UnderstandingCharacteristics}. The EAs used for these tasks -- leaving apart details -- very much follow the pseudo-code given in Algorithm~\ref{alg:ea_evolver}. In a nutshell the algorithm initializes a population $P$ with $\mu$ \emph{random uniform Euclidean}~(RUE) instances. I.~e., each of the $n$ city coordinates is placed uniformly at random within a bounding-box usually set to $[0,1]^2$ (see left-most plot in Figure~\ref{fig:mutation_operators}). Within the evolutionary loop one instance is selected to produce an offspring. The classical mutation operators either add some Gaussian noise to a subset of the points of the parent as illustrated in Figure~\ref{fig:mutation_operators}~(center) or re-locate a subset of the points by sampling new coordinates, again, uniformly at random. An elitist $(\mu+1)$-strategy assures that the best individuals with respect to the objective function survive. On termination the best instance is usually returned. It turned out that even though goal (D1) was successfully met, (D2) was not; in fact no difference to RUE instances was visible to the naked eye and as a consequence the feature space $\mathcal{F}$ was not well covered.
Later, this issue was approached by different groups of researchers. Gao et al.~\cite{GaoNN16feature} used a bi-level approach which aimed to evolve a feature-diverse population of instances with minimal performance quality requirements (see also the recent journal extension~\cite{doi:10.1162/evcoa00274}). To this end they developed a diversity measure based on a weighted sum approach to calculate each individuals contribution to the feature diversity in the population. Though appealing, the bi-level character delayed the search for low objective values. \changed{Later TSP instance generation was approached in the context of evolutionary diversity optimization~(EDO). In EDO, given a (close-to-)optimal solution OPT and a parameter $\alpha >1$ the goal is to evolve a population $P$ such that $f(I) \leq (1+\alpha)\cdot\text{OPT}$ for all $I \in P$ and a diversity measure $D : P \to \mathbb{R}$ is maximized. Promising contributions in this field adopted the star-discrepancy~\cite{neumann2018discrepancy} or well-known performance indicators from multi-objective optimization~\cite{neumann2019evolutionary} as feature-space diversity measure $D$.}
Bossek et al.~\cite{Bossek2019evolv} recently took a different path and introduced a sophisticated set of "creative" mutation operators. These operators addressed the short-comings of the aforementioned operators (re-location and Gaussian noise) by strongly affecting the parents' point coordinates. An easy to illustrate example is the \emph{implosion} mutation where all points within a random radius around a randomly sampled center of implosion are attracted towards the center (see right-most plot in Figure~\ref{fig:mutation_operators}). The authors showed in their study, adopting a $(5+1)$~EA where in each iteration one of these creative mutation operators was applied to the parent, that this approach -- applied many times to produce a set of instances -- covered a much wider part of the feature space without explicit feature-space diversity preservation at all.

All approaches discussed so far are very wasteful in the sense that only the best individual of the final population is returned as output or the whole usually small\footnote{Due to the computational costs of the instance generation process the population is usually small in order to allow for as many generations as possible.} final population for the approaches with diversity preservation. Nevertheless, many potentially interesting intermediate instances \added{become} extinct in the course of optimization. This is undesirable since the entire process is obviously very time consuming.\footnote{In recent studies on state-of-the-art inexact TSP solvers, for each generated instance the \underline{exact} optimum has to be calculated in order to calculate the PAR-score. E.~g., in~\cite{Bossek2019evolv} the generation of a single instance was given a time budget of 48h CPU time.} Hence, there is a need for a systematic means to memorize instances of interest.

\section{A Quality Diversity Approach}
\label{sec:method}

A straight-forward idea is to simply save all intermediate instances, i.~e., the footprint of the EA so to say. This approach is not target-oriented however since instances with large performance gaps are desirable; see diversity requirement (D1). We now present an idea that borrows ideas from quality diversity~(QD). QD is a recently emerged branch of evolutionary computation. In the so-called \emph{Map-elites} approach the \emph{target space} is partitioned into \emph{boxes} where each box stores the best solution found for all target space realizations closest to its box-center. This approach is highly successful, e.~g., in evolving different robot behavior~\cite{CullyM13} where the target space corresponds to the behavioral space of the robots.
In our setting the target space corresponds to the feature space $\mathcal{F} \subseteq \mathbb{R}^p$ we aim to cover. Each box of our QD-algorithm corresponds to one possible feature combination. The algorithm is designed for discrete features where, given a fixed instance size $n$, the number of possible realizations of each feature is countable and bounded. This is indeed the case for many TSP-features of interest, e.~g., $k$-NNG based features or depth-based MST features~\cite{Pihera2014MLTSP} which are frequently among the top discriminating features in AS-models~\cite{Pihera2014MLTSP,Kerschke2018leveraging,Seiler2020DeepLearning}.\footnote{E.~g., given $n$ cities, the number of weak connected components in the $k$-NNG can take values in $\{1, 2, \ldots, n-k\}$ only~\cite{HBPSTK2021tspnormalization}.} We refer the interested reader to recent work by Heins et al.~\cite{HBPSTK2021tspnormalization} who derived theoretical lower and upper bounds for a large collection of such features for feature normalization.
\begin{algorithm}[t]
	\SetKwInOut{Input}{input}
    \Input{Objective $f$, feature-mapping $F$.}
    Initialize empty map $M$\;
    Initialize instance $I$ at random\;
    Calculate $F(I)$ and store $I$ in $M[F(I)]$\;
    \While{Stopping condition not met} {
        Select an instance $I$ from $M$ uniformly at random\;
        Generate $I'$ by mutating $I$\;
        \If{Box $M[F(I')]$ is not covered}{
            Store $I'$ in $M[F(I')]$\;
        }\Else{
            Let $I''$ be the instance stored in box $M[F(I')]$\;
            \If{$f(I')$ is not worse than $f(I'')$}{
                Replace $I''$ with $I'$ in $M[F(I')]$\;
            }
        }
    }
    \Return{$M$}\;
    \caption{QD approach for instance generation.}
    \label{alg:qd_evolver}
\end{algorithm}
For each box our algorithm -- in classical QD-fashion -- stores the TSP instance found over time with the best performance value. More precisely, the algorithm proceeds as shown in Algorithm~\ref{alg:qd_evolver}. First, an empty map is initialized. The map takes the role of the "population" in Algorithm~\ref{alg:ea_evolver}. Next, an RUE instance $I$ is created at random and stored in the respective box of the map indexed by $I$'s feature-vector $F(I)$.\footnote{The map can be implemented as a hash-map allowing for very fast look-up operations.} From there on the following steps are performed until a stopping condition is met: an instance $I$ from the subset of \emph{covered} boxes, i.~e., those that are already assigned an instance, is sampled uniformly at random. Mutation is applied to $I$ which yields another instance $I'$. Next, if the box at $F(I')$ is empty (we say the box is \emph{hit for the first time} in the following), $I'$ is saved in any case regardless of its objective value. Otherwise, if the box is already covered, the objective function $f$ comes into play. Let $I''$ be the instance assigned to the box, i.~e., $F(I') = F(I'')$. If $f(I')$ is not worse than $f(I'')$ the box is \emph{updated} by replacing $I''$ with $I'$. The algorithm returns all instances stored in the map upon termination. Hence, the algorithm follows an elitist strategy in each box and thus box-wise the objective function is monotonically decreasing (w.~l.~o.~g., we assume minimization).
The approach is simple but nevertheless very generic and powerful and by no means restricted to the TSP domain that we adopt here for a proof-of-concept study in this contribution.

\section{Experiments on Simple Heuristics}
\label{sec:experiments_cheap}

In this and the next section we continue with an extensive empirical evaluation of the proposed approach. Our experimental setup is two-fold: In the \emph{cheap} setting we evolve small instances for simple TSP heuristics whereas in the \emph{expensive} setting we evolve larger instances with the goal to maximize performance differences between state-of-the-art heuristics. 
The cheap experiments -- due to fast evaluation of the objective function -- allow to explore the differences obtained after a huge number of iterations as a proof-of-concept. In the expensive setting we can afford only a fraction of the iterations due to computational limitations, but tackle interesting algorithms.

\subsection{Experimental setup}

In a first series of experiments we evolve small instances with $n=100$ nodes. We consider two insertion heuristics: \emph{Farthest-Insertion}~(FI) and \emph{Nearest-Insertion}~(NI)~\cite{Rosenkrantz2009tsp}. FI starts with a partial tour consisting of one randomly selected node. In $(n-1)$ iterations the remaining nodes are added until a TSP tour is constructed. In the selection phase the algorithm selects the node furthest away from the currently maintained sub-route. This node is injected into the existing partial route such that the least extension of the previous partial route is created. (NI) works very much alike with a slight difference in the selection phase: NI selects the nearest node. The goal is to evolve instances which span a two-dimensional feature space space nicely and minimize the tour length ratio $FI(I)/NI(I)$ or $NI(I)/FI(I)$ respectively. Ratios below~1 are desirable as this indicates a performance advantage for the "numerator" algorithm.
The feature combinations consist of combinatorial features calculated on basis of a MST of the instance or its $k$-NNG as introduced in~\cite{Pihera2014MLTSP}; for details we refer the reader to the original papers. More precisely, we consider feature combinations~(FCs) that were frequently selected by AS-models in various studies (see, e.~g., \cite{Pihera2014MLTSP,Bossek2019evolv}):
\begin{enumerate}
    \item[FC1)] Maximum size of a strong connected component in the 3-NNG (nng\_3\_strong\_components\_max) and the number of weak connected components in the 3-NNG~(nng\_3\_n\_weak).
    \item[FC2)] The number of strong connected components in the 5-NNG (nng\_5\_n\_strong) and the median depth of nodes in a MST of the input graph (mst\_depth\_median).
\end{enumerate} 
In the following we will make heavy use of the abbreviations FC1 and FC2 for ease of writing.
\changed{For each algorithm pair (FI vs. NI) and (NI vs. FI) and each of the two FCs we evolve instances by ten algorithms in total: QD implements Algorithm~\ref{alg:qd_evolver}. We compare QD against classic evolutionary algorithms $(1+1)$~EA and $(50+1)$~EA as well as EDO-focused EAs $(50+1)$~ED-IGD and $(50+1)$~EA-HV from~\cite{neumann2019evolutionary} which use inverted generational distance~(IGD) and the dimension-doubling hyper-volume indicator for (feature-)diversity maximization. These algorithms were among the best-performing EAs in~\cite{neumann2019evolutionary} (see the reference for details). EDO-based algorithms require a good initial population. Hence, we initialize each run with clones of a solution obtained by running $(1+1)$-EA for additional $10\,000$ iterations.
All algorithms are run with either only \emph{simple} mutation operators (re-location, Gaussian mutation) from Mersmann et al.~\cite{Mersmann13-TSPmeta} (suffix [simple]) or the full set of disruptive mutation operators introduced in Bossek et al.~\cite{Bossek2019evolv} (suffix [all]; see also Section~\ref{sec:problem}).\footnote{In each mutation step one of the mutation operators is selected uniformly at random.}
Note that the baseline algorithms return at most 50 instances and thus it is unfair to compare directly against these; the winner would be determined in advance -- QD -- since it stores all novel instances with not yet seen feature-combinations. Hence, for a fair comparison we modify the EAs to store instances the same way the proposed QD approach does (see Algorithm~\ref{alg:ea_with_archive_evolver}). We stress that this way of saving the footprint of the EAs has also not been done before in the literature.}
\begin{algorithm}[t]
	\SetKwInOut{Input}{input}
    \Input{Objective $f$, population size $\mu$, feature-mapping $F$.}
    Initialize empty map $M$\;
    Initialize population $P$ of $\mu$ instances at random\;
    Store $I \in P$ at $M[F(I)]$ for all $I \in P$\;
	\While{Stopping condition not met}{
        Select random instance $I \in P$\;
        Generate $I'$ by mutating $I$\;
        \If{Box $M[F(I')]$ is not covered}{
            Store $I'$ in $M[F(I')]$\;
        }\Else{
            Let $I''$ be the instance stored in box $M[F(I')]$\;
            \If{$f(I')$ is not worse than $f(I'')$}{
                Replace $I''$ with $I'$ in $M[F(I')]$\;
            }
        }
        \If{$f(I')$ is not worse than $f(I)$}{
            Replace $I$ with $I'$ in $P$\;
        }
    }
    \Return{$M$}\;
    \caption{$(\mu+1)$~EA (see Alg.~\ref{alg:ea_evolver}) with archive.}
    \label{alg:ea_with_archive_evolver}
\end{algorithm}
Each experiment was repeated \changed{30 times} with different random numbers generators seeds and a budget of $1\,000\,000$ function evaluations. Within each objective function evaluation the tour length of each TSP heuristic is calculated as the mean over 5 independent runs to account for stochasticity in the choice of the initial node for tour construction. All algorithms were implemented in the R programming language~\cite{Rlang}. Code and data will be publicly available in a GitHub repository upon acceptance.\footnote{GitHub repository: \url{https://github.com/jakobbossek/GECCO2022-QD-TSP}}

\subsection{Results}

We first study the number of boxes covered, i.~e., the number of instances / feature combinations hit by the algorithms over time. To this end we logged the load of the map $M$ over time.
\begin{figure}[t]
    \centering
    \includegraphics[width=0.6\columnwidth]{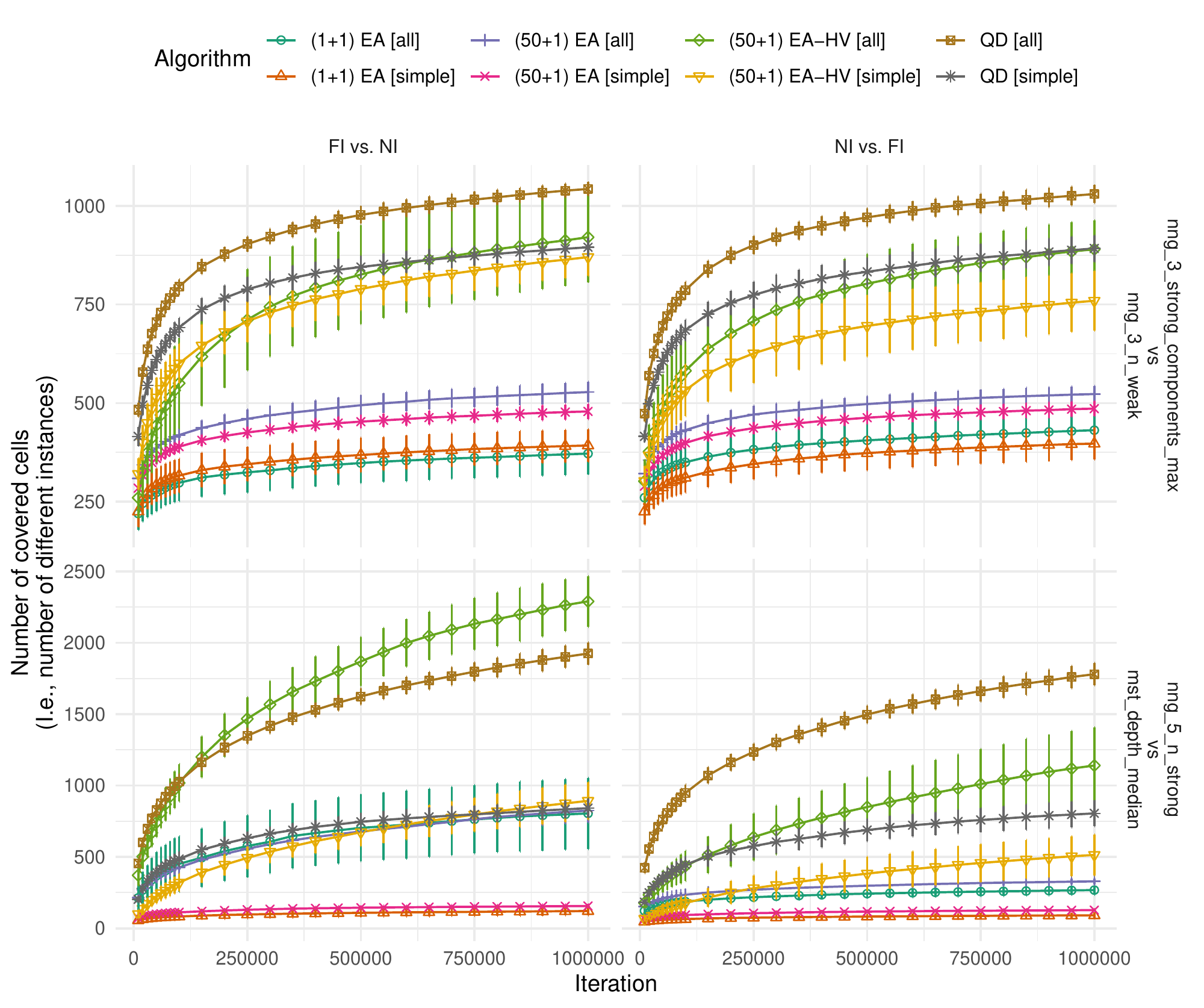}
    \caption{Mean number ($\pm$ standard deviation) of boxes covered (i.~e., instances detected) in the map over the course of the search for FC1 (top row) and FC2 (bottom row).}
    \label{fig:nr_of_instances_over_time}
\end{figure}

Figure~\ref{fig:nr_of_instances_over_time} shows the progress of the mean number of boxes covered (with one standard-deviation error bars). \changed{Note that we show $(50+1)$~EA-IGD in all subsequent plots as a representative EDO-EA since it performs very much like the EA-HV version. We observe that both QD approaches are highly superior to the classic EA-baselines with respect to feature space coverage. After termination the mean number of boxes covered by QD~[all] surpasses the best classic EAs' value by a factor of $>2$ for FC1 and up to a factor of $>4$ for FC2 and the algorithm pair (NI vs. FI).
The EDO-EAs perform much better than their classic counterpart. However, expect for (FI. vs. NI) on FC2 they produce less instances than QD~[all] and QD-algorithms show a high robustness / very low variation. Table~\ref{tab:statistics_cheap} shows the mean and standard deviation of boxes covered among other statistics that will be discussed later. The observations are confirmed. A Kruskal-Wallis test with Bonferroni-correction rejects the zero hypothesis $H_0: \text{median}(X) > \text{median}($QD~[all]$)$ for every competitor evolver $X$ at significance level $\alpha=0.0001$ for all settings expect for the mentioned (FI. vs. NI) on setting FC2; the results are highly significant.}
Another important observation is that QD~[all] dominates the whole field of competitors significantly in the first few thousand iterations. This is important for computationally more involved instance generation task (see Section~\ref{sec:experiments_expensive}).
As expected, all algorithms tend to perform much better
\begin{table*}
\renewcommand{\tabcolsep}{3pt}
\renewcommand{\arraystretch}{0.8}
\begin{scriptsize}
\caption{\label{tab:statistics_cheap}Table of different aggregated values of interest for the experiments on the simple insertion heuristics. Best values (for the statistics of covered  or objective values) or maximal values (for box update statistics) are \colorbox{gray!40}{highlighted}.}
\centering
\begin{tabular}[t]{llrrrrrrrrrrrrrr}
\toprule
\multicolumn{2}{c}{ } & \multicolumn{7}{c}{\textbf{FI vs. NI}} & \multicolumn{7}{c}{\textbf{NI vs. FI}} \\
\cmidrule(l{3pt}r{3pt}){3-9} \cmidrule(l{3pt}r{3pt}){10-16}
\multicolumn{2}{c}{ } & \multicolumn{3}{c}{\textbf{Nr. of cells}} & \multicolumn{2}{c}{\textbf{Cell statistics}} & \multicolumn{2}{c}{\textbf{Objective}} & \multicolumn{3}{c}{\textbf{Nr. of cells}} & \multicolumn{2}{c}{\textbf{Cell statistics}} & \multicolumn{2}{c}{\textbf{Objective}} \\
\cmidrule(l{3pt}r{3pt}){3-5} \cmidrule(l{3pt}r{3pt}){6-7} \cmidrule(l{3pt}r{3pt}){8-9} \cmidrule(l{3pt}r{3pt}){10-12} \cmidrule(l{3pt}r{3pt}){13-14} \cmidrule(l{3pt}r{3pt}){15-16}
\textbf{FC} & \textbf{Algorithm} & \textbf{mean} & \textbf{std} & <\textbf{med} & \textbf{upd.} & \textbf{hits} & \textbf{best} & \textbf{median} & \textbf{mean} & \textbf{std} & <\textbf{med} & \textbf{upd.} & \textbf{hits} & \textbf{best} & \textbf{median}\\
\midrule
 & (1+1) EA [all] & 371.93 & 52.20 & 180.00 & \cellcolor{gray!40}{\textbf{165.00}} & 186079.00 & \cellcolor{gray!40}{\textbf{0.53}} & 0.68 & 431.33 & 56.13 & 215.00 & 45.00 & 192185.00 & \cellcolor{gray!40}{\textbf{0.77}} & 0.97\\

 & (1+1) EA [simple] & 392.53 & 41.07 & 193.00 & 70.00 & \cellcolor{gray!40}{\textbf{487405.00}} & 0.65 & 0.77 & 397.13 & 40.60 & 199.00 & \cellcolor{gray!40}{\textbf{45.00}} & \cellcolor{gray!40}{\textbf{430152.00}} & 0.87 & 1.00\\

 & (50+1) EA [all] & 527.90 & 25.96 & 263.00 & 76.00 & 30129.00 & 0.55 & 0.70 & 522.90 & 20.88 & 260.00 & 27.00 & 30250.00 & 0.83 & 0.98\\

 & (50+1) EA [simple] & 478.83 & 19.39 & 240.00 & 31.00 & 51435.00 & 0.69 & 0.77 & 485.90 & \cellcolor{gray!40}{\textbf{19.60}} & 243.00 & 26.00 & 51992.00 & 0.88 & 1.00\\

 & (50+1) EA-HV [all] & 920.43 & 113.77 & 475.00 & 25.00 & 12855.00 & 0.57 & 0.71 & 889.77 & 73.86 & 444.00 & 19.00 & 11843.00 & 0.87 & 0.98\\

 & (50+1) EA-HV [simple] & 870.03 & 47.61 & 434.00 & 21.00 & 10710.00 & 0.71 & 0.78 & 759.33 & 75.32 & 381.00 & 20.00 & 13330.00 & 0.90 & 0.99\\

 & (50+1) EA-IGD [all] & 976.50 & 90.07 & 503.00 & 27.00 & 20658.00 & 0.63 & 0.73 & 882.87 & 74.75 & 447.50 & 19.00 & 19466.00 & 0.84 & 0.98\\

 & (50+1) EA-IGD [simple] & 865.10 & 44.73 & 437.50 & 21.00 & 17756.00 & 0.71 & 0.78 & 739.80 & 69.00 & 369.50 & 22.00 & 18890.00 & 0.91 & 1.00\\

 & QD [all] & \cellcolor{gray!40}{\textbf{1042.73}} & \cellcolor{gray!40}{\textbf{17.50}} & \cellcolor{gray!40}{\textbf{518.00}} & 72.00 & 4610.00 & 0.56 & \cellcolor{gray!40}{\textbf{0.64}} & \cellcolor{gray!40}{\textbf{1029.80}} & 22.67 & \cellcolor{gray!40}{\textbf{516.00}} & 32.00 & 4433.00 & 0.80 & \cellcolor{gray!40}{\textbf{0.93}}\\

\multirow{-10}{*}{\raggedright\arraybackslash FC1} & QD [simple] & 894.90 & 23.87 & 445.50 & 38.00 & 3981.00 & 0.66 & 0.73 & 891.93 & 33.72 & 446.00 & 27.00 & 4531.00 & 0.87 & 0.96\\
\cmidrule{1-16}
 & (1+1) EA [all] & 804.43 & 248.08 & 432.00 & \cellcolor{gray!40}{\textbf{100.00}} & 116604.00 & \cellcolor{gray!40}{\textbf{0.53}} & 0.64 & 267.77 & 32.19 & 133.00 & 39.00 & 310009.00 & \cellcolor{gray!40}{\textbf{0.78}} & 1.02\\

 & (1+1) EA [simple] & 120.83 & 30.30 & 57.00 & 90.00 & \cellcolor{gray!40}{\textbf{582995.00}} & 0.62 & 0.76 & 91.07 & 19.40 & 44.00 & \cellcolor{gray!40}{\textbf{95.00}} & \cellcolor{gray!40}{\textbf{700120.00}} & 0.81 & 1.02\\

 & (50+1) EA [all] & 824.33 & 84.32 & 402.50 & 44.00 & 34771.00 & 0.55 & 0.64 & 328.80 & 21.66 & 164.00 & 26.00 & 84605.00 & 0.83 & 0.99\\

 & (50+1) EA [simple] & 156.03 & \cellcolor{gray!40}{\textbf{13.86}} & 78.00 & 38.00 & 246284.00 & 0.66 & 0.77 & 127.43 & \cellcolor{gray!40}{\textbf{8.83}} & 64.00 & 34.00 & 244907.00 & 0.88 & 1.02\\

 & (50+1) EA-HV [all] & 2289.90 & 177.68 & 1148.50 & 24.00 & 8302.00 & 0.59 & 0.69 & 1140.17 & 265.98 & 565.00 & 21.00 & 26899.00 & 0.85 & 0.99\\

 & (50+1) EA-HV [simple] & 892.73 & 130.89 & 462.00 & 22.00 & 31931.00 & 0.69 & 0.76 & 513.83 & 143.77 & 269.00 & 21.00 & 85790.00 & 0.89 & 1.00\\

 & (50+1) EA-IGD [all] & \cellcolor{gray!40}{\textbf{2335.60}} & 162.63 & \cellcolor{gray!40}{\textbf{1184.00}} & 28.00 & 20174.00 & 0.58 & 0.70 & 1047.53 & 233.71 & 505.00 & 17.00 & 49680.00 & 0.87 & 1.00\\

 & (50+1) EA-IGD [simple] & 768.37 & 121.46 & 387.50 & 23.00 & 153460.00 & 0.68 & 0.77 & 412.63 & 123.33 & 196.00 & 23.00 & 302422.00 & 0.90 & 1.01\\

 & QD [all] & 1924.33 & 76.97 & 969.00 & 51.00 & 3315.00 & 0.55 & \cellcolor{gray!40}{\textbf{0.61}} & \cellcolor{gray!40}{\textbf{1779.77}} & 79.11 & \cellcolor{gray!40}{\textbf{886.00}} & 27.00 & 5240.00 & 0.81 & \cellcolor{gray!40}{\textbf{0.97}}\\

\multirow{-10}{*}{\raggedright\arraybackslash FC2} & QD [simple] & 841.63 & 67.22 & 419.00 & 40.00 & 11698.00 & 0.64 & 0.71 & 804.87 & 90.07 & 397.50 & 27.00 & 15761.00 & 0.88 & 0.99\\
\bottomrule
\end{tabular}
\end{scriptsize}
\end{table*}
\begin{figure*}
    \centering
    \includegraphics[width=0.48\columnwidth]{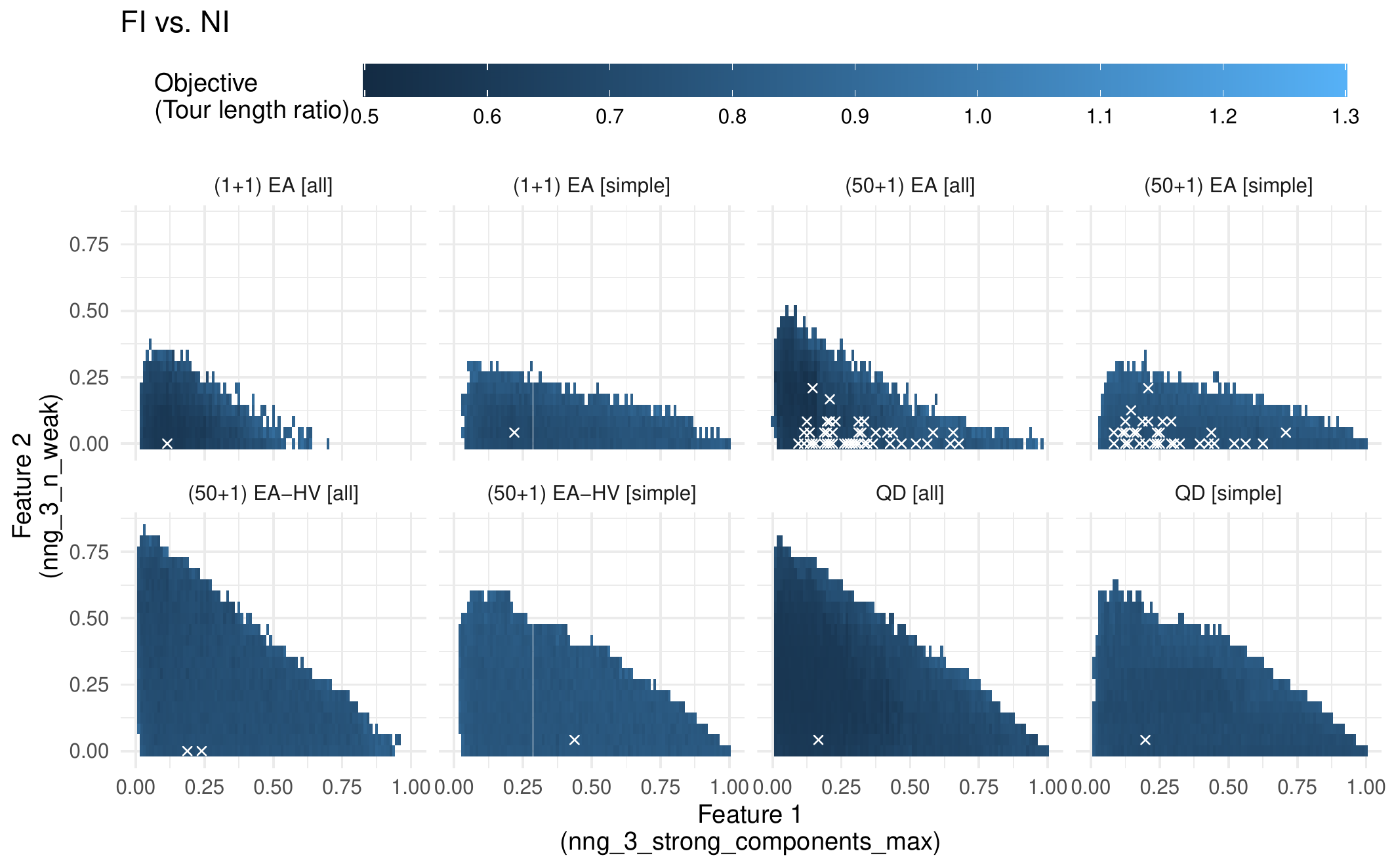}
    \includegraphics[width=0.48\columnwidth]{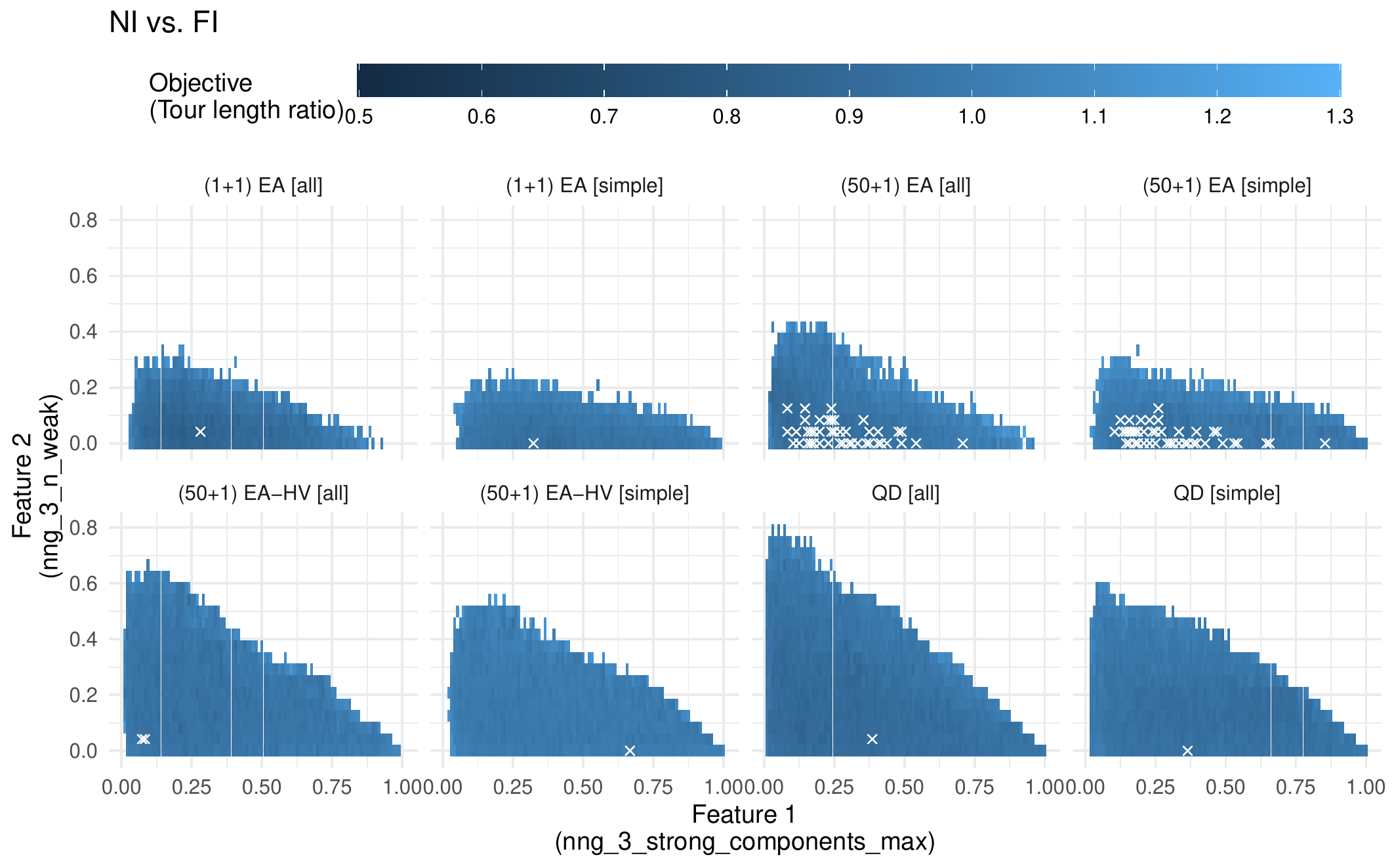}
    \includegraphics[width=0.48\columnwidth, trim={0 0 0 30pt}, clip]{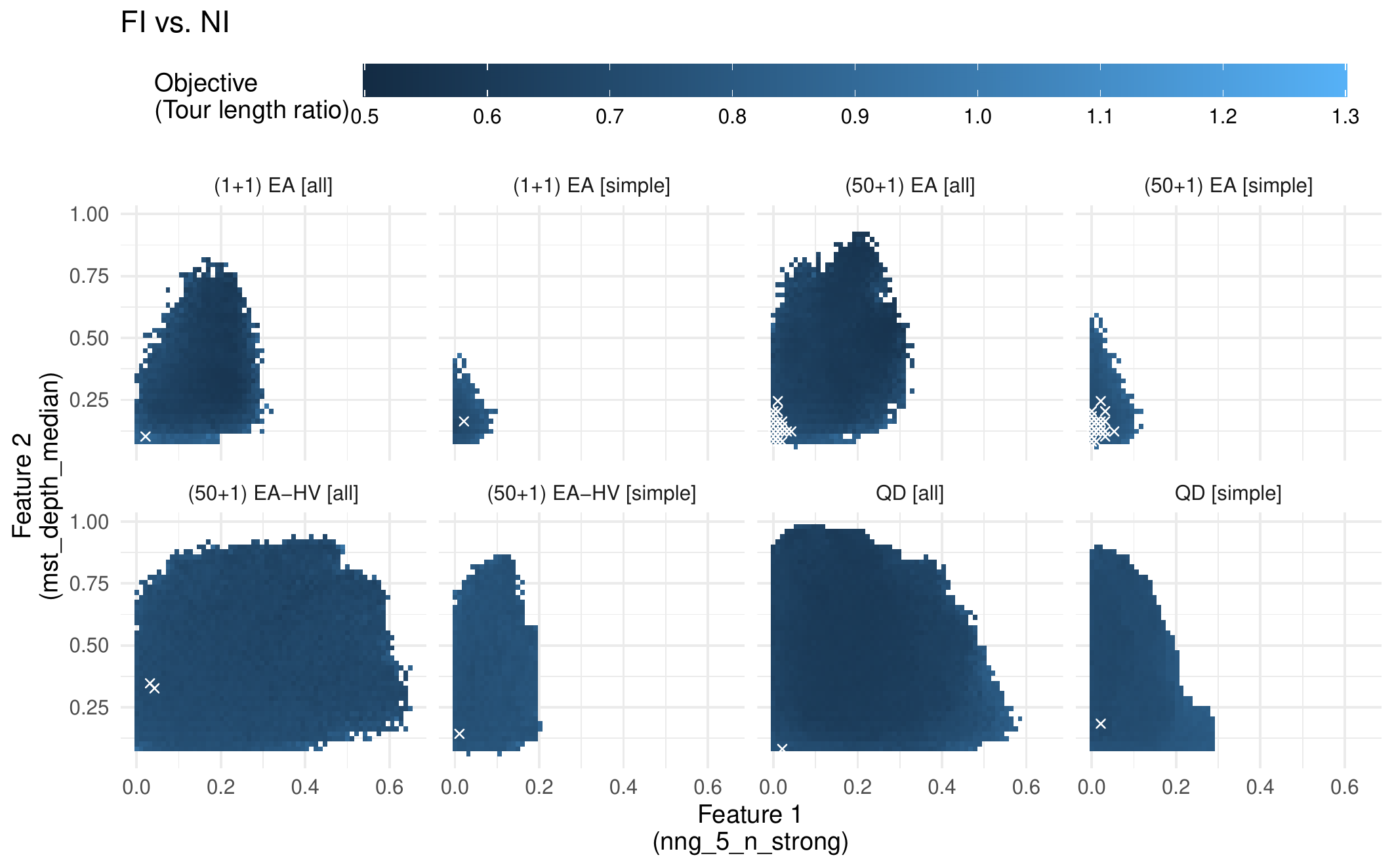}
    \includegraphics[width=0.48\columnwidth, trim={0 0 0 30pt}, clip]{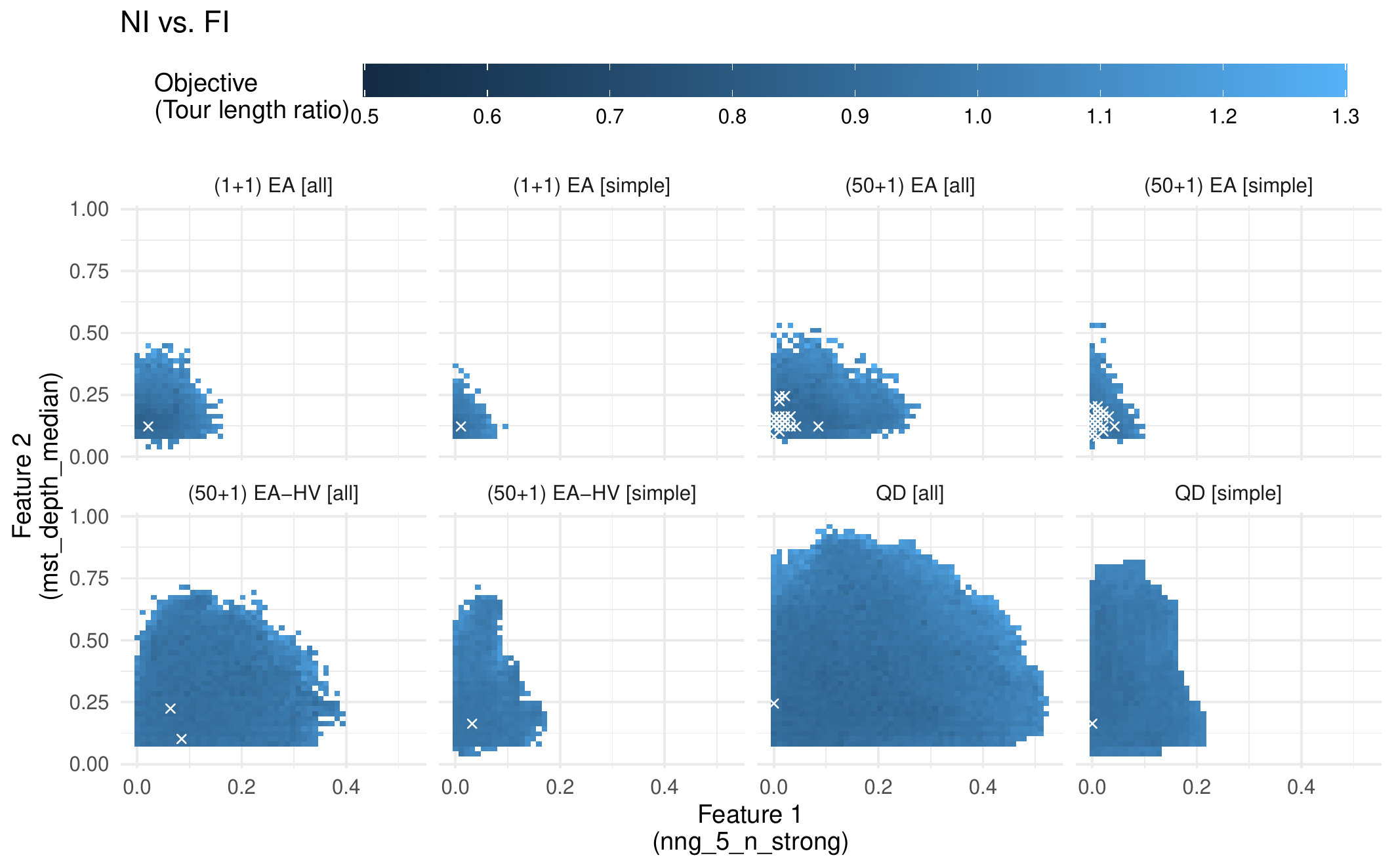}
    \vspace{-0.25cm}
    \caption{Representative feature space coverage obtained by the different algorithms for feature combination FC1 (top row) and FC2 (bottom row). White crosses mark the boxes hit by the initial population for (FI vs. NI; left) and (NI vs. FI; right). The tiles share a common continuous color gradient with the same limits to make the differences in the objective values visible.}
    \label{fig:feature_space}
\end{figure*}
if all mutation operators are used due to the disruptive nature of the creative mutation~\cite{Bossek2019evolv}. However, the benefit of disruptive over simple mutation is surprisingly low for FC1 and very strong for FC2. 
Figure~\ref{fig:feature_space} shows the feature space (normalized to $[0,1]$~\cite{HBPSTK2021tspnormalization}) coverage by means of exemplary but representative runs for FC1 and FC2.
The covered boxes are colored by their objective value (smaller is better) and white crosses represent the boxes hit by the initial solutions. We see a slight benefit for the *[all] algorithms and FC1. However, the feature space coverage is indeed very similar. Explanation: consider the spread of initial solutions in Figure~\ref{fig:feature_space} for the $(50+1)$~EA variants on FC1. Recall that each initial solution is generated by placing $n$ points uniformly at random within $[0,1]^2$ (RUE; see Section~\ref{sec:problem}). These initial instances already show a nice distribution of this particular feature space indicating that simple mutation operators (which produce again RUE-like mutants) are sufficient to further explore the FC1-space. The FC2 space coverage in Figure~\ref{fig:feature_space} shows a different picture. Here, the initial population is very close to the origin. I.~e., for all initial solutions the feature vectors in FC2-space are quite similar. This is plausible since an RUE instance is likely to have a low number of strong connected components in the $5$-NNG and likewise a rather small median depth of nodes in an MST. In the FC2-space, the strong mutation of the *[all] variants allows all algorithms to explore the feature space much more thoroughly. 

\begin{figure}
    \centering
    \includegraphics[width=0.75\columnwidth]{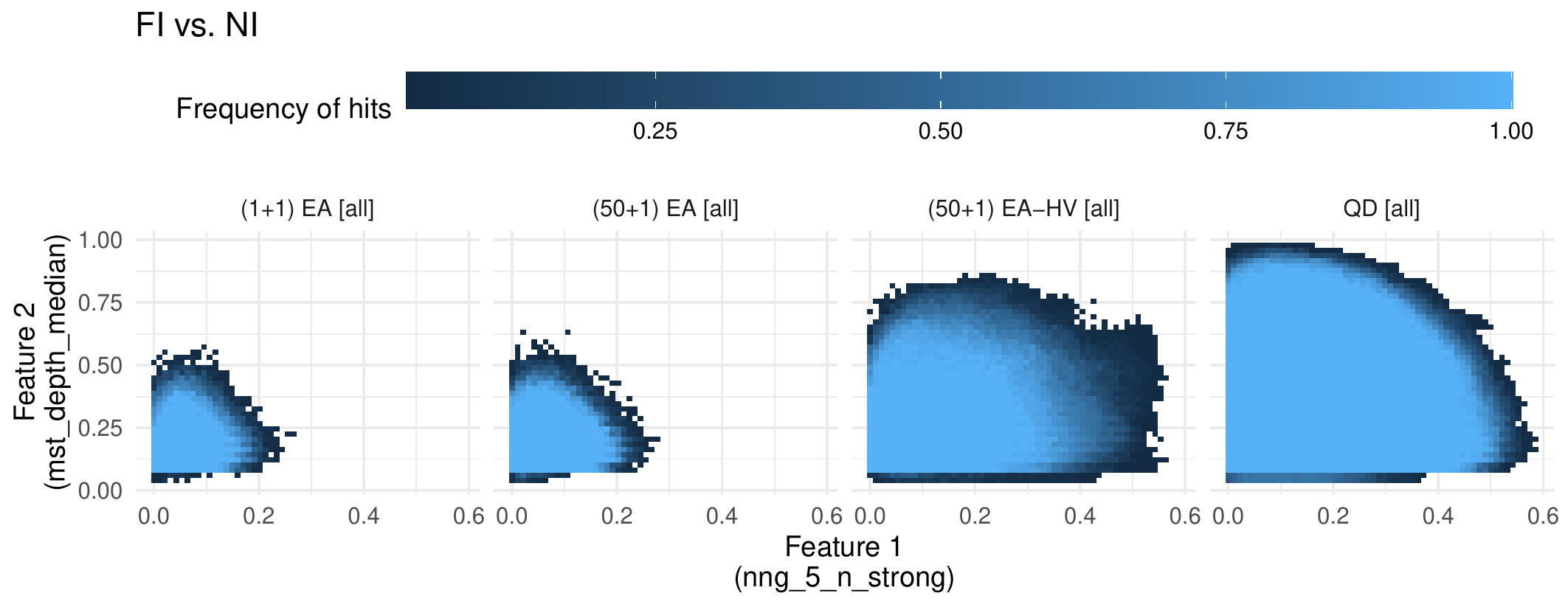}
    \caption{Distribution of the coverage frequency across all 30 independent runs for (NI vs. FI) on FC2.}
    \label{fig:feature_space_frequency}
\end{figure}

\begin{figure}
    \centering
    \includegraphics[width=0.75\columnwidth]{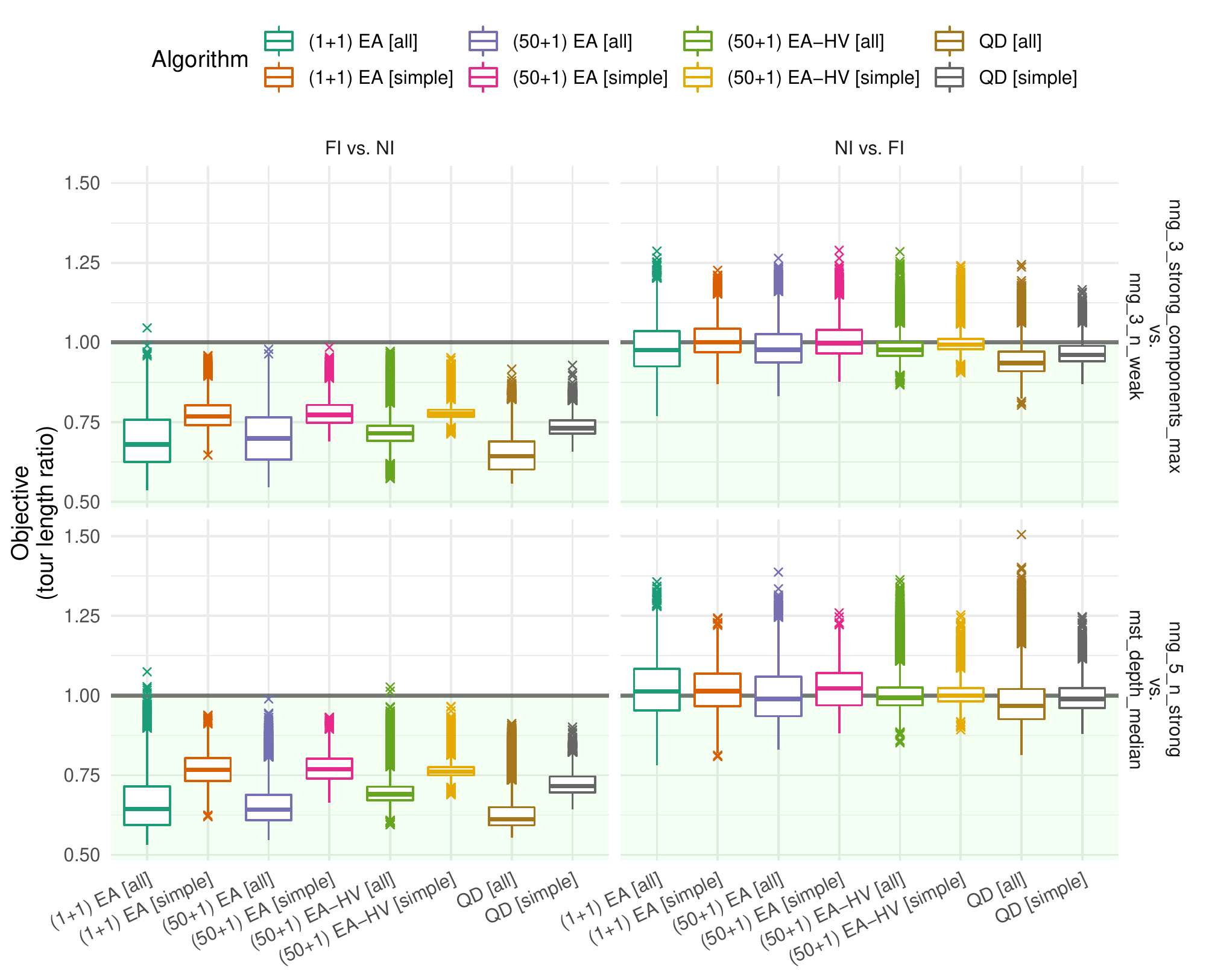}
    \caption{Distribution of the objective values (ratio of tour lengths) of all instances over all independent runs. Data is split by the direction of the optimization process (columns) and the feature space covered (rows).}
    \label{fig:boxplot_objective_simple_heuristics}
\end{figure}

Comparing the *[simple] and *[all] plots in Figure~\ref{fig:feature_space} for FC2 and also the respective progress over time in Figure~\ref{fig:nr_of_instances_over_time} gives rise to another question: why do the \added{classic and EDO-based} $(\mu+1)$~EA~[all] variants cover much more of the FC2 feature space when the objective is to evolve instances which are easier to solve for FI than for NI? And why is this not the case for their *[simple] counterparts? One might think that this has to be a bug since an instance is always saved if its feature vector was not seen so far; this \emph{first hit} of a box is independent of the objective value. However, the reason lies in the way the survival selection of the EAs works: the mutant replaces the parent if and only if its objective value is not worse than that of the parent~(see Algorithm~\ref{alg:ea_with_archive_evolver}) \added{or -- for the EDO-EAs -- diversity maximization takes place only if the instance satisfies the minimum quality constraint}. Since it is much more challenging to produce instances which are harder for FI than for NI (see Figure~\ref{fig:boxplot_objective_simple_heuristics} where the objective values barely fall below~1) this is an unlikely event to happen. Consequently, in particular $(1+1)$~EA suffers from few updates of its "population" and its only population member serves as the template for the vast majority of the produced offspring. A single mutation (regardless of the type: explosion, implosion etc.) might be insufficient in producing a previously unseen feature combination. This lack of variation leads to the poor performance on (NI vs. FI). In the reverse case, (FI vs. NI), things are different and the population members get updated more often which allows for a more thorough exploration of the feature space since the "source material" changes more frequently. This phenomenon is not observed for FC1. As pointed out already the FC1-space can be easily explored even by simple mutation. 
Recall that the identified issue may be a major hindrance for the classical $(\mu+1)$ -- and in particular the $(1+1)$-approaches -- from the literature to improve the objective since it carries over directly to EAs without footprint archive.
\added{We close the discussion of box-coverage by investigating the robustness of the approach. Figure~\ref{fig:feature_space_frequency} colors boxes by the coverage frequency across all runs. QD explores the feature space much more reliably with little variation. The observations are the same for all other settings.}

\begin{figure}
    \centering
    \includegraphics[width=0.75\columnwidth]{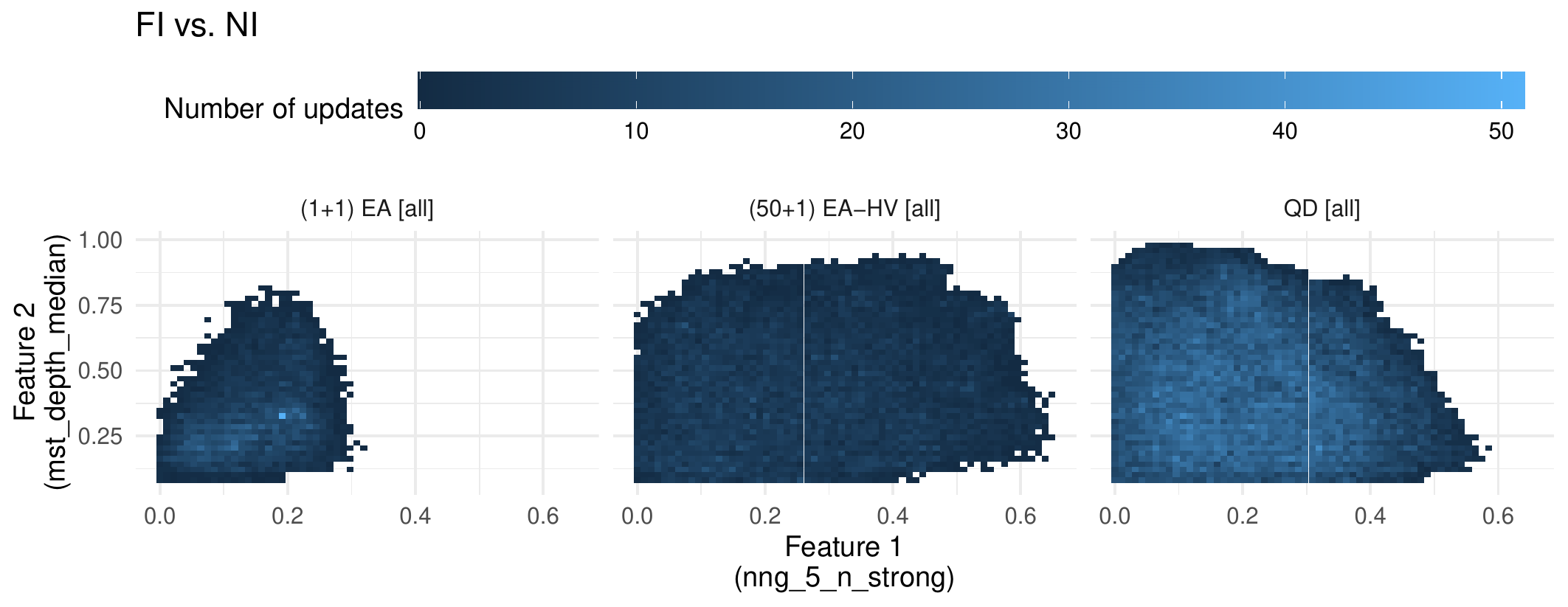}
    \includegraphics[width=0.75\columnwidth]{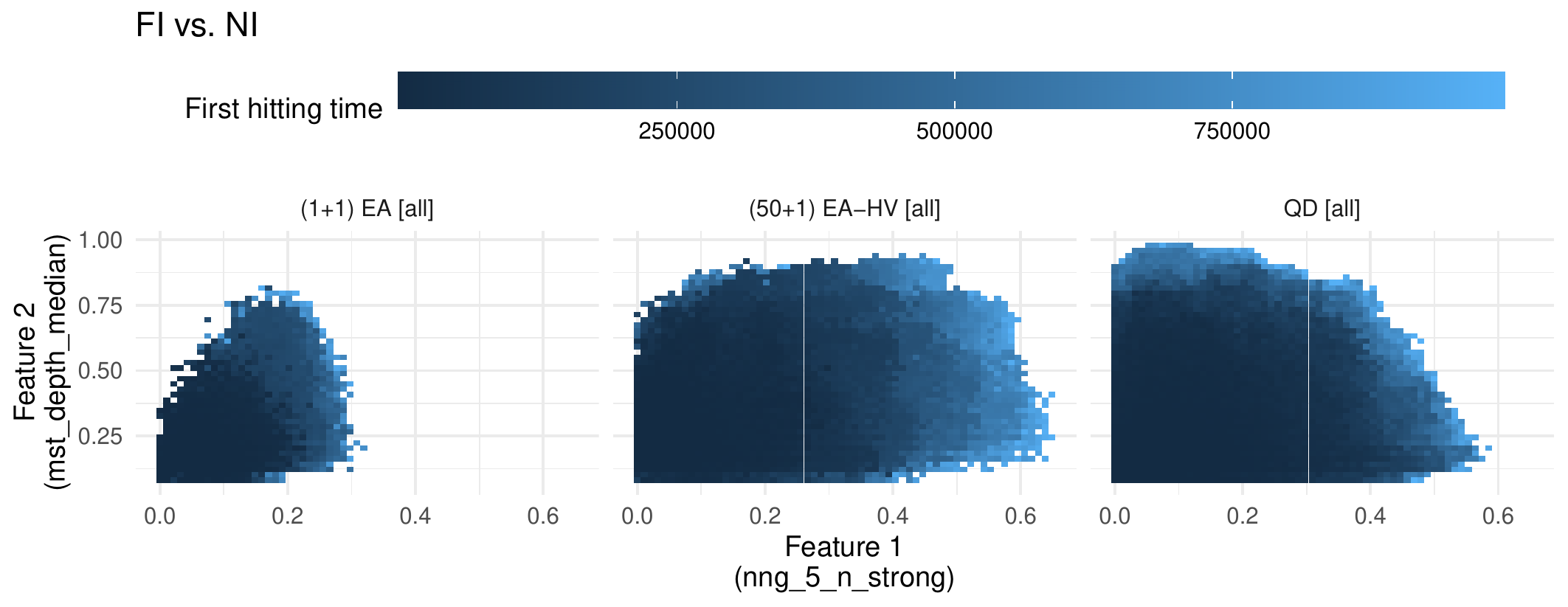}
    \caption{FC2 with boxes colored by the number of updates, i.~e., the number of times the instance stored in the box was replaced with an instance not worse with respect to the objective score~(top row). In the bottom row boxes are colored by the first hitting time / (first discovery time).}
    \label{fig:FI_NI_nng5_mst_tile_nupdates_firsthit}
\end{figure}

So far we investigated the ability of the algorithms to cover the feature space, one of the two goals of the evolving procedure. We now turn the focus towards the second \added{equally if not more important} goal: minimizing the tour length ratio. Figure~\ref{fig:boxplot_objective_simple_heuristics} shows the distribution of the objective values on all instances over all runs produced by the algorithms. The panels split the data by feature combination (rows) and objective "direction" ((FI vs. NI) and (NI vs. FI); columns). Recall that the algorithms strive to minimize the tour length ratio. Thus, values in the green shaded region below~1 are desirable. We observe that QD~[all] dominates the field for all combinations of features and target directions with respect to the median objective value. In particular in the (NI vs. FI) setting, where finding instances which are harder to solve for NI seems difficult, the median objective values are clearly located in the greenish area whereas the median objective scores of the competitor algorithms are closer or even above~1. With the exception of (NI vs. FI) on FC2, the upper quartile for QD~[all] is even lower than the median of the other algorithms. This observation highlights the need for stronger exploration of the feature space and shows the superiority of the QD-approach.
\added{Note also that the median objective values produced by the EDO-EAs are worse than the respective QD-values. This is particularly visible in (FI. vs. NI) setting and both FC1 and FC2. Recall that QD~[all] covered less boxes in the FC2 scenario than the EDO-EAs. However, it shines when it comes to the median objective values. This is also confirmed by the precise numbers in Table~\ref{tab:statistics_cheap}.}
\begin{figure}
    \centering
    \includegraphics[width=\columnwidth]{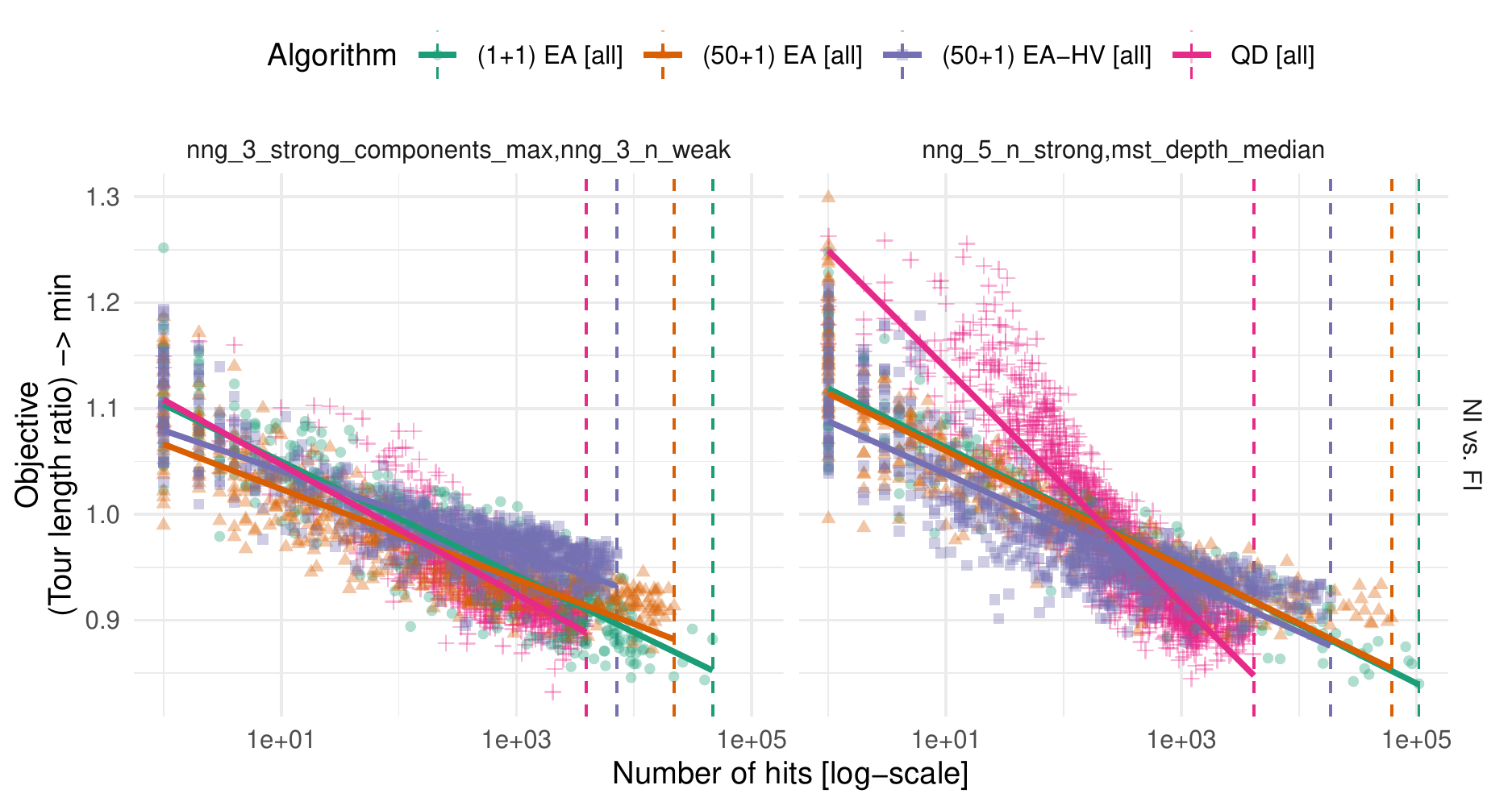}
    \caption{Scatter-plot of the number of box-hits on log-scale and the objective value of the instance stored in the box. The plot shows the first run by example.}
    \label{fig:FI_NI_scatterplot_nhits_vs_obj}
\end{figure}
Keep in mind that the data used for Figure~\ref{fig:boxplot_objective_simple_heuristics} is the union of all instances of all \changed{30} algorithm runs. Considering the advantage of QD~[all] with respect to feature space coverage the QD approaches in consequence produced significantly more easy/hard instances; a clear benefit.
Regarding the overall "best" objective values, in most cases the $(1+1)$~EAs dominates. This is in line with our expectation. Recall that (1) the $(1+1)$~EAs explores a significantly smaller share of the feature space and (2) has the same budget of function evaluations. Hence, the search process is much more focused and boxes are thus hit much more often. 
This increases the chances to update respective boxes with better instances. Figure~\ref{fig:FI_NI_scatterplot_nhits_vs_obj} shows the relation between the (log-scaled) number of hits on the abscissa and the objective value of the instance stored in the respective boxes after termination on the ordinate. The plots shows the result of one particular run of the algorithms. However, the patterns are the same for all other runs. We see obvious negative linear relationships. In addition, we see that for a subset of the feature combinations in the $(\mu+1)$~EAs the number of hits surpasses the maximum number of hits for the respective QD algorithm by orders of magnitude~(recall that the abscissa is on logarithmic scale).
\added{An interesting observation is that the number of box-hits of the EDO-EAs is also much higher. Nevertheless, QD remains the winner with respect to the objective. The reason is the survival selection of the EDO-EAs which may lead to frequent box-hits which do not improve upon the objective however.}
Figure~\ref{fig:FI_NI_nng5_mst_tile_nupdates_firsthit} in the top row shows this narrowed search by means of example in the FC2-space explored by $(1+1)$~EA~[all], \added{$(50+1)$~EA-HV~[all]} and QD~[all] where boxes are colored by the number of actual updates. While for QD~[all] the updates are distributed across the whole explored feature space (leave apart the border regions) only a few boxes are updated more that 30 times in the $(1+1)$~EA~[all] run. \added{The $(50+1)$~EA-HV~[all] updates cells less often which explains its rather poor performance with respect to the objective values discussed before.}
The \emph{box statistics} in Table~\ref{tab:statistics_cheap} support these observations. They show the maximum number of updates and hits of boxes over all runs. $(1+1)$~EA shows a huge maximum of hits up to $\approx 52\%$ of the total number of iterations on (NI vs. FI) and FC2. The values for QD in comparison are minuscule.
The decreased number of updates at the borders can be explained by the hardness to produce instances with such "extreme" feature vectors. The bottom row in Figure~\ref{fig:FI_NI_nng5_mst_tile_nupdates_firsthit} shows the same data colored by the iteration number of the boxes' first hit. Some of the boxes were only discovered after more than $800\,000$ iterations. This observation together with the increasing trend in Figure~\ref{fig:nr_of_instances_over_time} indicates that even longer runs would hit even more boxes.

\begin{figure}
    \centering
    \includegraphics[width=0.75\columnwidth]{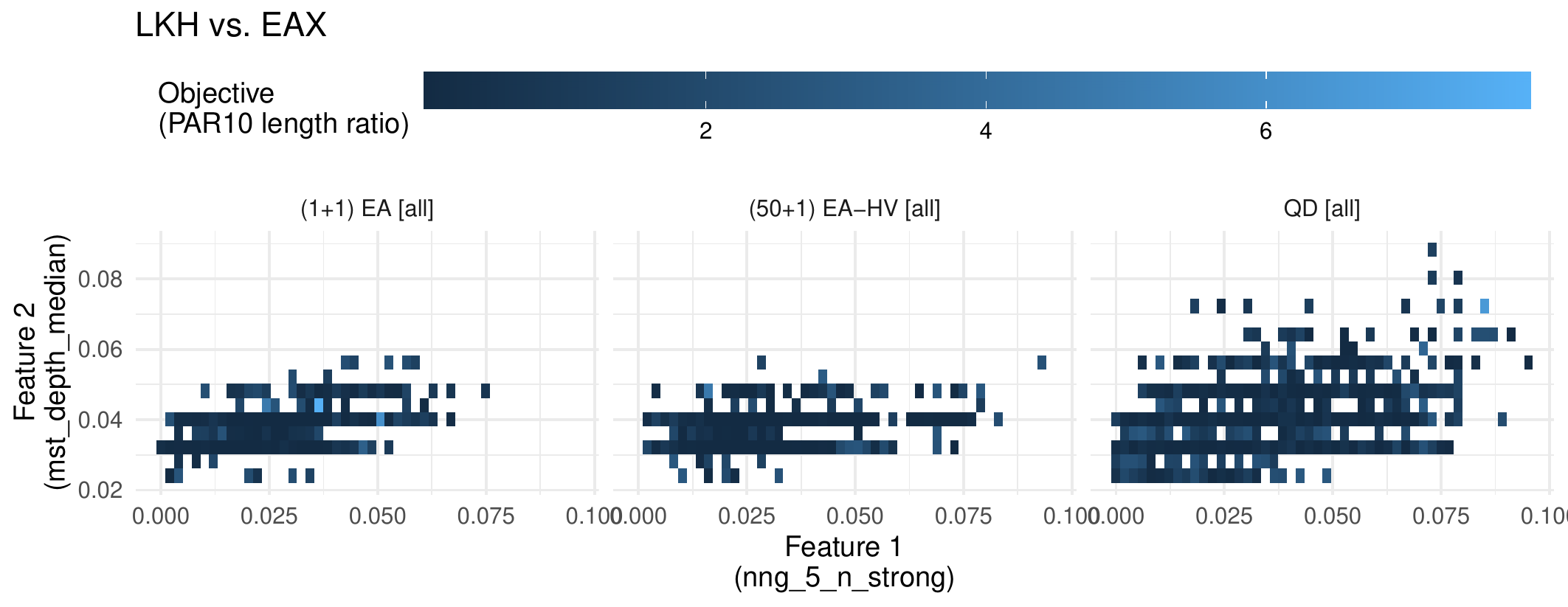}
    \caption{Representative feature space coverage obtained by the different algorithms for feature combination FC2 and state-of-the-art heuristics LKH and EAX.}
    \label{fig:feature_space_expensive}
\end{figure}
\section{Experiments on State-of-the-Art Inexact TSP Solvers}
\label{sec:experiments_expensive}

We now consider EAX and LKH and the feature combination FC2 only since the results in Section~\ref{sec:experiments_cheap} indicate that this space is harder to cover. Both algorithms are state-of-the-art in inexact TSP solving. EAX~\cite{Nagata1997EAX,Nagata2013powerful} is a sophisticated genetic algorithm based on the edge-assembly-crossover operator and entropy population diversity preservation. LKH~\cite{Helsgaun2009submoves} is a local-search algorithm whose main driver is the Lin-Kernighan heuristics for $k$-opt moves. Both algorithms are capable of solving even large instances to optimality in reasonable time. Complementary behavior on a wide range of instances motivated algorithm selection studies~\cite{Kotthoff2015improving,kerschke2018,Bossek2019evolv,HBPSTK2021tspnormalization}.

\subsection{Experimental setup}
Preliminary experiments show that EDO-based algorithms perform worse than QD in this computationally demanding domain with respect to both box-coverage and objective value since the number of iterations is severely reduced to $\approx 2\,000$. 
\begin{table*}
\renewcommand{\tabcolsep}{4pt}
\renewcommand{\arraystretch}{0.8}
\centering
\begin{scriptsize}
\caption{\label{tab:statistics_expensive}Table of aggregated values of interest for the experiments on state-of-the-art TSP solvers EAX and LKH. Best values (for the statistics of covered or objective values) or maximal values (for box update statistics) are \colorbox{gray!40}{highlighted}.}
\centering
\begin{tabular}[t]{llrrrrrrrrrrrrrr}
\toprule
\multicolumn{2}{c}{ } & \multicolumn{7}{c}{\textbf{EAX vs. LKH}} & \multicolumn{7}{c}{\textbf{LKH vs. EAX}} \\
\cmidrule(l{3pt}r{3pt}){3-9} \cmidrule(l{3pt}r{3pt}){10-16}
\multicolumn{2}{c}{ } & \multicolumn{3}{c}{\textbf{Nr. of boxes}} & \multicolumn{2}{c}{\textbf{Box statistics}} & \multicolumn{2}{c}{\textbf{Objective}} & \multicolumn{3}{c}{\textbf{Nr. of boxes}} & \multicolumn{2}{c}{\textbf{Box statistics}} & \multicolumn{2}{c}{\textbf{Objective}} \\
\cmidrule(l{3pt}r{3pt}){3-5} \cmidrule(l{3pt}r{3pt}){6-7} \cmidrule(l{3pt}r{3pt}){8-9} \cmidrule(l{3pt}r{3pt}){10-12} \cmidrule(l{3pt}r{3pt}){13-14} \cmidrule(l{3pt}r{3pt}){15-16}
\textbf{FC} & \textbf{Algorithm} & \textbf{mean} & \textbf{std} & <\textbf{med} & \textbf{upd.} & \textbf{hits} & \textbf{best} & \textbf{median} & \textbf{mean} & \textbf{std} & <\textbf{med} & \textbf{upd.} & \textbf{hits} & \textbf{best} & \textbf{median}\\
\midrule
 & (1+1) EA [all] & 390.83 & 100.32 & 199.50 & 7.00 & 141.00 & \cellcolor{gray!40}{\textbf{0.00}} & \cellcolor{gray!40}{\textbf{0.82}} & 574.38 & 153.52 & 292.00 & 8.00 & 368.00 & \cellcolor{gray!40}{\textbf{0.02}} & \cellcolor{gray!40}{\textbf{0.33}}\\

 & (50+1) EA-HV [all] & 442.83 & 240.29 & 247.50 & \cellcolor{gray!40}{\textbf{7.00}} & \cellcolor{gray!40}{\textbf{277.00}} & 0.00 & 0.99 & 712.00 & 242.13 & 321.00 & \cellcolor{gray!40}{\textbf{8.00}} & \cellcolor{gray!40}{\textbf{387.00}} & 0.03 & 0.34\\

\multirow{-3}{*}{\raggedright\arraybackslash FC1} & QD [all] & \cellcolor{gray!40}{\textbf{877.50}} & \cellcolor{gray!40}{\textbf{54.98}} & \cellcolor{gray!40}{\textbf{430.50}} & 6.00 & 24.00 & 0.00 & 0.85 & \cellcolor{gray!40}{\textbf{882.57}} & \cellcolor{gray!40}{\textbf{130.68}} & \cellcolor{gray!40}{\textbf{441.00}} & 7.00 & 25.00 & 0.03 & 0.36\\
\cmidrule{1-16}
 & (1+1) EA [all] & 128.57 & 23.75 & 65.00 & 8.00 & 237.00 & 0.00 & 0.38 & 135.38 & \cellcolor{gray!40}{\textbf{12.81}} & 67.50 & 11.00 & \cellcolor{gray!40}{\textbf{942.00}} & \cellcolor{gray!40}{\textbf{0.03}} & \cellcolor{gray!40}{\textbf{0.29}}\\

 & (50+1) EA-HV [all] & 116.60 & \cellcolor{gray!40}{\textbf{22.94}} & 59.00 & 8.00 & \cellcolor{gray!40}{\textbf{371.00}} & \cellcolor{gray!40}{\textbf{0.00}} & \cellcolor{gray!40}{\textbf{0.30}} & 137.88 & 24.19 & 73.00 & \cellcolor{gray!40}{\textbf{11.00}} & 797.00 & 0.03 & 0.31\\

\multirow{-3}{*}{\raggedright\arraybackslash FC2} & QD [all] & \cellcolor{gray!40}{\textbf{281.00}} & 62.02 & \cellcolor{gray!40}{\textbf{141.00}} & \cellcolor{gray!40}{\textbf{8.00}} & 44.00 & 0.00 & 0.42 & \cellcolor{gray!40}{\textbf{353.67}} & 60.70 & \cellcolor{gray!40}{\textbf{174.00}} & 9.00 & 72.00 & 0.04 & 0.38\\
\bottomrule
\end{tabular}
\end{scriptsize}
\end{table*}
\begin{figure}
    \centering
    \includegraphics[width=0.49\columnwidth]{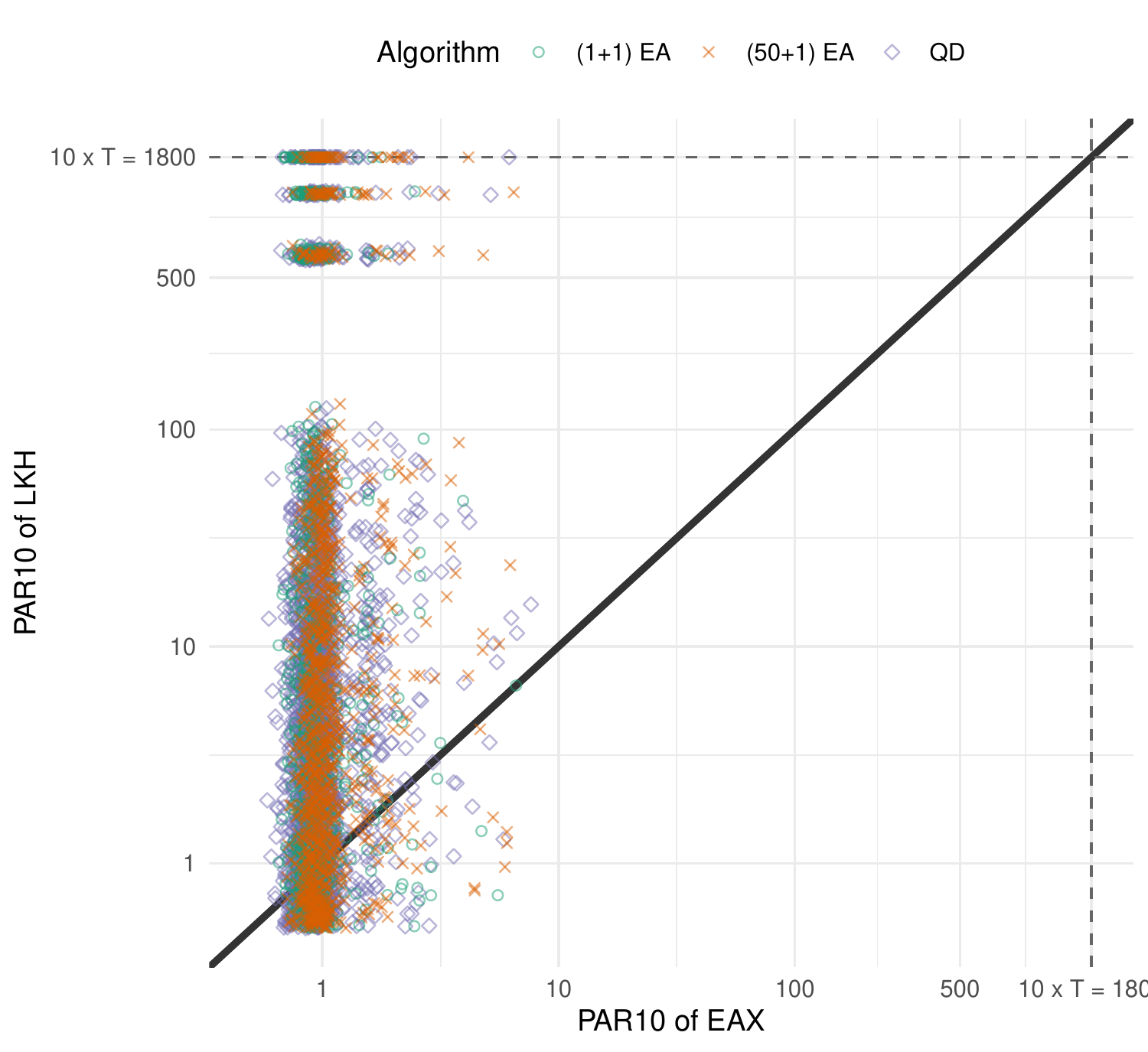}
    \includegraphics[width=0.49\columnwidth]{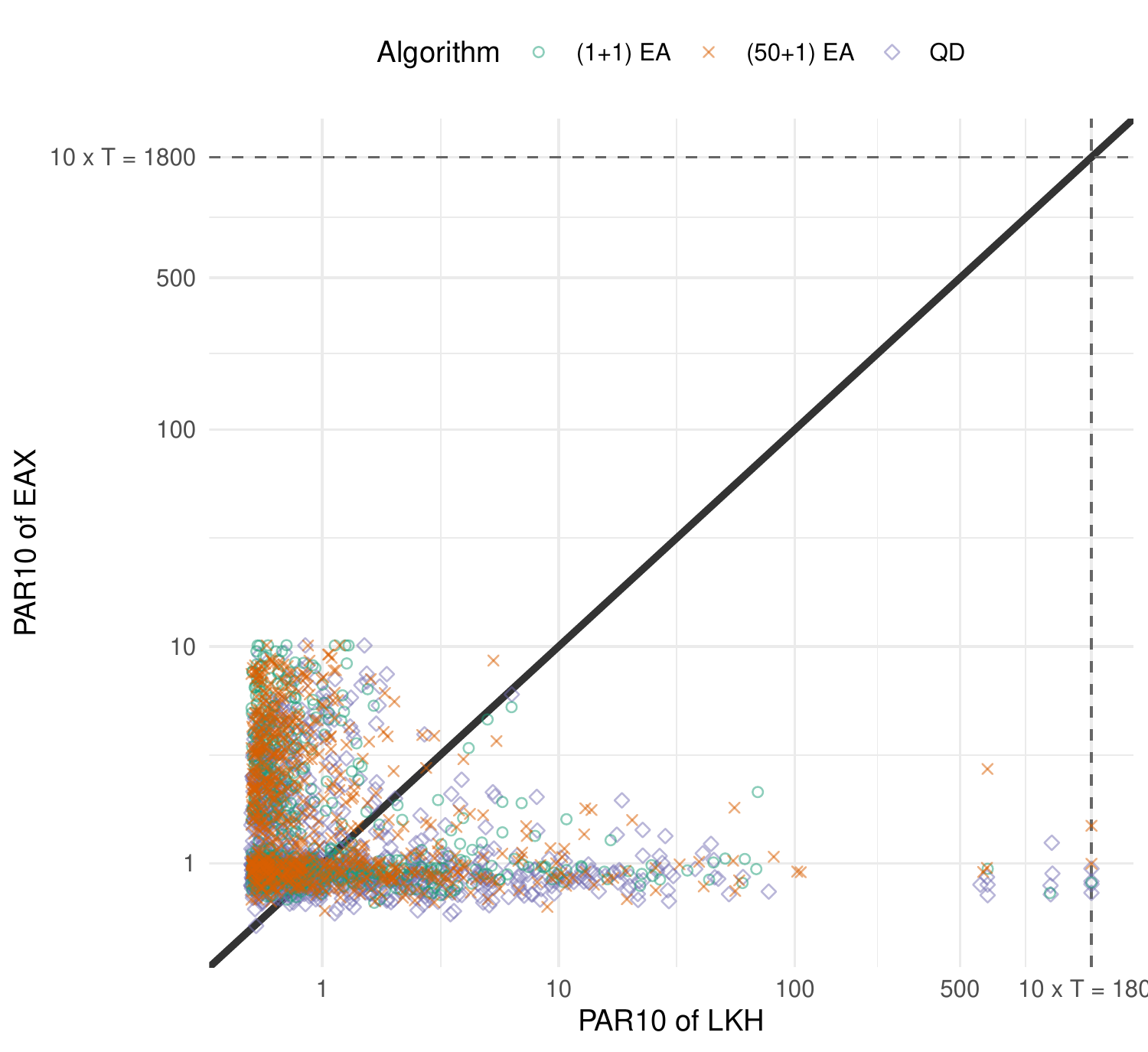}
    \caption{Scatter-plots of PAR10 scores of the two competing algorithms EAX and LKH. Points are colored by the evolving algorithm. The left plot shows the results for the experiments with the goal to produce instances which are easy for EAX and hard for LKH; the right plot shows the vice versa. Points on the diagonal line indicate equal performance of both algorithms on the respective instance. Points above/below the diagonal indicate a runtime-advantages for one of the algorithms. The gray dashed lines indicate the maximum possible PAR10-score of $10 \cdot 180$~[s].}
    \label{fig:scatter_par_scores}
\end{figure}
This is in-line with observations extracted from Figure~\ref{fig:nr_of_instances_over_time} and Figure~\ref{fig:boxplot_objective_simple_heuristics}. Hence, in the following we use $(1+1)$~EA, $(50+1)$~EA and QD.
In line with \cite{Bossek2019evolv} we evolve instances with $n=500$ nodes and minimize the ratio of the penalized average runtime~(PAR10~\cite{bischl2016}) scores, the mean of the runtimes of $R$ runs of $A$ on $I$ where \emph{failure runs} are penalized with $10 \cdot T$, $T$ being the cutoff-time; a run is termed a \emph{failure} if the algorithm does not find the optimum within time $T$. The objective function evaluation requires a single run of the exact TSP solver \texttt{concorde}~\cite{Applegate2009TSP} to calculate an optimal tour. We set $R=3$ and $T=180~[s]$ and run each evolving algorithm with 48h wall-time.

\subsection{Results}
The observations on feature-space coverage in Figure~\ref{fig:feature_space_expensive} are in par with Figure~\ref{fig:feature_space} (bottom row) in the cheap setting. However, due to the costly objective function evaluation the algorithms performs significantly less iterations which explains the lower total coverage. QD~[all] dominates the other algorithms with respect to the covered area in all scenarios due to only up to $\approx$~2000 performed generations. Table~\ref{tab:statistics_expensive} shows summary statistics for the number of covered boxes, the objective values and box update/hit statistics. The numbers confirm the good performance of all approaches with respect to the objective values and the advantage of the QD approaches with respect to feature space coverage. One observation worth to mention is that the mean number of boxes covered is higher for all algorithms for (LKH vs. EAX) in comparison to (EAX vs. LKH). The reason is that LKH hits the cutoff-time of $180$ seconds frequently while this is rarely the case for EAX. Put differently, it is easier for the algorithms to evolve an instance where LKH hits the time limit than the vice verse. Figure~\ref{fig:scatter_par_scores} shows this nicely by plotting the PAR10-scores of EAX and LKH against each other for the evolving algorithms $(1+1)$~EA~[all], $(50+1)$~EA-HV~[all] and QD~[all]. The three clusters in the top left corner of the left plot indicate that on many evolved instances LKH has either one, two or even three failure runs. In contrast, EAX times our rarely if the goal is to produce EAX-hard instances. 

\section{Conclusion}
\label{sec:conclusion}

The generation of TSP problem instances which are diverse with respect to solver performance and instance features is of utmost importance, e.~g., in the field of algorithm selection. We introduced a flexible quality diversity~(QD) approach to evolve instances. Instances are mapped to their feature vectors. In an iterative evolutionary process not yet seen instances / feature combinations are stored in any case; instances are overwritten if newly generated instances share the same feature combination, but show a stronger performance difference for two competing TSP algorithms.
We compared our approach to classical $(\mu+1)$~EAs and versions adapted to feature-diversity optimization from the literature modified to store their footprints in a similar way. The results show impressively the capability of the QD approach in covering a wide range of feature combinations -- and thus producing much more instances in a single, less wasteful run -- while competing or even outperforming the classical approaches with respect to objective scores. These results motivate a broad avenue of future work. Here, we only name a few of our most promising ideas: application to $>2$ features and other domains, introducing selection bias towards border regions, or storing multiple instances per box.

\section{Acknowledgements}
J. Bossek acknowledges support by the European Research Center for Information Systems (ERCIS). This work has been supported by the Australian Research Council (ARC) through grants DP190103894 and FT200100536. The authors thank the anonymous reviewers for their immensely valuable feedback.

\bibliographystyle{unsrt}  
\bibliography{bib}

\end{document}